\pdfoutput=1

\documentclass[11pt]{article}

\usepackage[]{emnlp2021}

\usepackage{times}
\usepackage{graphicx}
\usepackage{tablefootnote}
\usepackage{latexsym}

\usepackage[T1]{fontenc}

\usepackage[utf8]{inputenc}

\usepackage{microtype}
\usepackage{verbatim}

\title{Team Enigma at ArgMining-EMNLP 2021: \\ Leveraging Pre-trained Language Models for Key Point Matching}

\author{
    Manav Nitin Kapadnis\footnotemark[1]$\thanks{~~Equal contribution.}$\hspace{1.5mm}, Sohan Patnaik\footnotemark[1] \hspace{0mm}, \\ \textbf{Siba Smarak Panigrahi}\footnotemark[1], \hspace{0.5mm} \textbf{Varun Madhavan}\footnotemark[1],\hspace{1.mm} \textbf{Abhilash Nandy}\\ Indian Institute of Technology Kharagpur\\

    \small{\texttt{\{iammanavk, sohanpatnaik106, sibasmarak.p, varun.m.iitkgp, nandyabhilash\}@gmail.com}}
}

\begin{document}
\maketitle
\begin{abstract}
We present the system description for our submission towards the Key Point Analysis Shared Task at ArgMining 2021. Track 1 of the shared task requires participants to develop methods to predict the match score between each pair of arguments and keypoints, provided they belong to the same topic under the same stance. We leveraged existing state of the art pre-trained language models along with incorporating additional data and features extracted from the inputs (topics, key points, and arguments) to improve performance. We were able to achieve \emph{mAP} strict and \emph{mAP} relaxed score of 0.872 and 0.966 respectively in the evaluation phase, securing 5th place\footnote{All results and leaderboard standings are reported using the default evaluation method (explained in section \ref{sec:resultsdiscussions})} on the leaderboard. In the post evaluation phase, we achieved a \emph{mAP} strict and \emph{mAP} relaxed score of 0.921 and 0.982 respectively. All the codes to generate reproducible results on our models are available on Github\footnote{\url{https://github.com/manavkapadnis/Enigma_ArgMining}}.

\end{abstract}
\vspace{-0.4cm}
\section{Introduction} 
\label{sec:introduction}
\vspace{-0.2cm}

The Quantitative Summarization - Key Point Analysis (KPA) Shared Task requires participants to identify the keypoints in a given corpus. Formally, given an input corpus of relatively short, opinionated texts focused on a particular topic, KPA aims to identify the most prominent keypoints in the corpus. Hence the goal is to condense free-form text into a set of concise bullet points using a well-defined quantitative framework.
In track 1, given a debatable topic, a set of keypoints per stance, and a set of crowd arguments supporting or contesting the topic, participants must report for each argument the corresponding match score for each keypoint under the same stance towards the topic. 
In track 2, we are required to build a language model that would generate keypoints given a set of arguments and a topic and finally find the match score of that particular keypoint with the argument. We mainly focused on the first track.

We frame the task of identifying the most prominent keypoints as a sentence similarity task, obtaining the most similar keypoints corresponding to a given argument. 



\section{Related Work} 
\label{sec:relatedwork}
\vspace{-0.2cm}

Sentence similarity is gaining much attention in the research community due to its versatility in various natural language applications such as text summarization \citep{similarity_for_text_summarization}, question answering \citep{ashok-etal-2020-simsterq}, sentiment analysis \citep{9415802} and plagarisim detection \citep{lo-simard-2019-fully}. 
Two major approaches to quantitatively measure similarity have been proposed - 

\begin{itemize}
  \item \textbf{Lexical similarity}, as the name suggests, is a measure of the extent or degree of lexicon overlap between two given sentences, ignoring the semantics of the lexicons. 
  \item \textbf{Semantic similarity} takes into account the meaning or semantics of the sentences. Deep Learning based approaches are typically leveraged to create dense representations of sentences, which are then compared using statistical methods like cosine similarity. 
\end{itemize}

Since the \emph{ArgKP-2021} dataset \citep{revise1add} contains crowd arguments for or against a particular stance, naturally, we expect some paraphrasing in the arguments put forth by different people. This indicates that semantic similarity would be an appropriate measure of similarity. However, we observe the problem of semantic drift \citep{jansen-2018-multi} in keypoint - argument pairs. Hence, we add additional lexical overlap and syntactic parse based features to improve performance (details on the features can be found in Section \ref{sec:methodology}).

\section{Dataset Description} 
\label{sec:dataset}
\vspace{-0.2cm}


The \emph{ArgKP-2021} dataset \citep{revise1add} which was the main dataset used for the shared task consists of approximately 27,520 argument/keypoint pairs for 31 controversial topics. Each of the pairs is labeled as matching or non-matching, along with a stance towards the topic. The train data comprises of 5583 arguments and 207 keypoints, the validation data comprises of 932 arguments and 36 keypoints and the test data comprises of 723 arguments and 33 keypoints. \\
Additionally, since external datasets were permitted, we experimented with two more datasets i.e., the IBM Rank 30k dataset \citep{gretz2019largescale} and the Semantic Textual Similarity or STS dataset \citep{cer2017semeval} (described in section \ref{subsection:existing-datasets}) to train our model before fine-tuning on the \emph{ArgKP-2021} dataset. The \emph{STS} dataset comprises of 8020 pairs of sentences, whereas the IBM Rank 30k dataset comprises of 30497 pairs of arguments and keypoints. 
\section{Implementation Details} 
\label{sec:methodology}
\vspace{-0.2cm}


In this section, we elaborate on our experiments and methodology to find the best-performing models. The section is organized to describe the addition of dependency parsing features in Section \ref{subsection:dep-parse-feat}, parts of speech features in Section \ref{subsection:parts-of-speech-feat}, Tf-idf features in Section \ref{subsection:tf-idf-feat}, and the use of external datasets in Section \ref{subsection:existing-datasets}.

\subsection{Baseline Transformer Model Architecture}
\label{subsec:baseline-architecture}


In recent work, Transformer \citep{vaswani2017attention} based pre-trained language models like BERT \citep{devlin-etal-2019-bert}, RoBERTa \citep{liu2019roberta}, BART \citep{lewis2019bart}, and DeBERTa \citep{he2021deberta}, have proven to be very powerful in learning robust context-based representations of lexicons and applying these to achieve state of the art performance on a variety of downstream tasks. 

We leverage these models for learning contextual representations of a keypoint - argument pair. The keypoints and arguments are individually concatenated, along with the topic (in the same order) for additional context information. We then obtain the contextual representation of this triplet and concatenate to it an encoded feature vector of additional features (one of Dependency Parse based features, Parts-of-Speech based features, and Tf-idf vectors). This concatenated vector was then passed through dense layers and a sigmoid activation to get a final similarity score in the desired range of $[0, 1]$, as shown in Figure \ref{fig:transformer-dep-feat}.


\subsection{Dependency Parsing Features}
\label{subsection:dep-parse-feat}

To capture the syntactic structure of the sentences, we added the dependency parse tree of the sentence as an additional feature. 

To obtain the same, we used the open-source tool \emph{spacy} \footnote{\url{https://spacy.io/}}. 
The dependency features are then label encoded according to descending order of occurrences. Consider three unique dependency features in all the concatenated sentences of the original dataset, namely, `aux', `amod', and `nsubj'. Let `aux', `nsubj', and `amod' be the descending order of count in the dataset, then `aux' is encoded as one, `nsubj' as two and 'amod' is encoded as three. All the names of unique features can be found in the supplementary material.

\begin{figure}[h!]
    \centering
  \includegraphics[scale = 0.40]{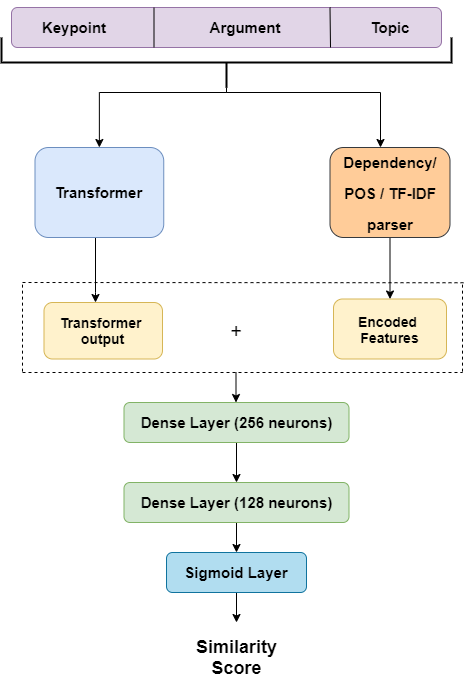}
  \caption{Model Architecture ("+" implies concatenation)}
  
  \label{fig:transformer-dep-feat}
\end{figure}

These encoded dependency features are then concatenated to the output of the transformer model and passed to subsequent layers as shown in Figure \ref{fig:transformer-dep-feat}.

\subsection{Parts of Speech Features}
\label{subsection:parts-of-speech-feat}
With a similar motive as before, i.e., to better capture the syntactic structure of the sentences, we experimented with Part-Of-Speech (POS) Features as well. 

As before, we used the open-source tool \emph{Spacy} to obtain POS labels for each lexicon, which were then label encoded according to descending order of occurrences. The encoded feature vector is then concatenated to the output of the transformer model and fed to the subsequent layers.

\subsection{Tf-idf features}
\label{subsection:tf-idf-feat}

In addition to semantic overlap, we wished to see if adding lexical overlap-based features would improve the ability of the model to identify similar sentences. To this end, we obtained the Tf-idf vector of the (keypoint, argument, topic) triplet (with padding). As before, the encoded feature vector is then concatenated to the output of the transformer model and fed further to the subsequent layers.


\subsection{External Datasets}
\label{subsection:existing-datasets}
We further tried to experiment with sentence similarity pre-training task on two additional datasets. The two datasets used were the STS benchmark dataset and the IBM Debater® - IBM Rank 30k dataset.

For the STS dataset, we normalized the target similarity score to bring the scores between $0$ and $1$. No additional preprocessing was done to the text. The two input sentences were concatenated into a single sentence and then directly fed to the model. We trained our model on STS dataset for 6 epochs and on the main dataset for 3 epochs. 

For the IBM Rank 30k dataset, we used the MACE \citep{hovy-etal-2013-learning} Probability score as the target column, which signifies the argument quality score for the corresponding topic. This is analogous to our approach for main task, wherein we output a similarity score for each argument-keypoint pair. No preprocessing was done to the text, the argument and topic were concatenated into a single sentence and then fed to the model. We trained our model on the IBM Rank 30k dataset for 3 epochs and on the main dataset for 3 epochs. 

Due to resource constraints, we were not able to perform pre-training on both the additional datasets one after another.
\section{Results and Discussions} 
\label{sec:resultsdiscussions}
\vspace{-0.2cm}
 
After we had concluded our experiments, a new evaluation method was proposed by organizers, which removes the positive bias towards a system that predict less true positives in high confidence. In the default evaluation metric a perfect recall is attained only when all positive ground truth labels are predicted, whereas the new method allows a perfect recall score when the top 50\% of the predictions (ranked by confidence) are positive. However, since we had completed all our experiments at this point, it was not feasible to rerun all our experiments in the given time frame. Hence we have reported all our results according to the default evaluation method.


Among all the transformer models without the use of external datasets, we found BART-large to perform best, along with DeBERTa-large with Tf-idf as additional features, achieving the best \emph{mAP} strict and \emph{mAP} relaxed score of 0.909, 0.982 and 0.911, 0.987 respectively. All the reported results are averaged over three seeds.

Table \ref{tab: table of vanilla text results} describes our experiments with different Transformer-based contextual language models without using any additional features. Recent improvements to the state-of-the-art in contextual language models in BART and DeBERTa perform significantly better than BERT. Further, BART is pre-trained using various self-supervised objectives such as token masking, sentence permutation, document rotation, token deletion and text infilling, unlike other models that mostly use either masked language modelling or next sentence prediction. In our opinion, the tasks of sentence permutation and document rotation help the model get a better understanding of context at the sentence level, and thus, are helpful when considering the keypoint matching task. We also observe that the \emph{large} version of the models, trained on more data with more parameters, perform significantly better than the \emph{base} versions, as expected.

\begin{table}[h!]
\centering
\resizebox{\columnwidth}{!}{\begin{tabular}{ccc}
\hline
\textbf{Model}      & \textbf{mAP Strict}     & \textbf{mAP Relaxed}    \\
\hline
BERT-base           & 0.804 $\pm$ 0.037          & 0.910 $\pm$ 0.050          \\
RoBERTa-base        & 0.826 $\pm$ 0.051          & 0.930 $\pm$ 0.032          \\
BART-base           & 0.824 $\pm$ 0.030          & 0.908 $\pm$ 0.020          \\
DeBERTa-base        & 0.894 $\pm$ 0.020          & 0.973 $\pm$ 0.015          \\
BERT-large          & 0.821 $\pm$ 0.025          & 0.924 $\pm$ 0.006          \\
RoBERTa-large       & 0.892 $\pm$ 0.003          & 0.970 $\pm$ 0.015          \\
BART-large & \textbf{0.909 $\pm$ 0.011} & \textbf{0.982 $\pm$ 0.003} \\
DeBERTa-large       & 0.889 $\pm$ 0.030          & 0.979 $\pm$ 0.010   \\
\hline
\end{tabular}}
\caption{Results of Transformer models}
\label{tab: table of vanilla text results}
\end{table}


Table \ref{tab:add_features_table} shows the best performing results obtained by concatenating one of the following - Dependency Parse features, POS features, and Tf-idf features. We note that out of the three feature vectors methods, Tf-idf features performs the best. Tf-idf gives a relation/measure of lexical overlap between the argument and keypoint, whereas the other features (POS and Dependency Parse) just expand on the sentence structures of the argument and the keypoint, without expressing the relation between the same. Thus it is observed that Tf-idf performs better than the other two feature vectors. In table \ref{tab:add_features_table}, we report the best-performing transformer-based models for each feature vector. Detailed results (each transformer model with each feature) can be found in the Appendix which is present in the supplementary material. We could not perform combination of all the syntactic features due to limited GPU memory availability.

\begin{table}[h!]
\centering
\resizebox{\columnwidth}{!}{\begin{tabular}{cccc}
\hline
\textbf{Feature} & \textbf{Best Model} & \textbf{mAP Strict} & \textbf{mAP Relaxed} \\
\hline
Dep\tablefootnote{Encoded dependency features (section \ref{subsection:dep-parse-feat})}    & BART-large    & 0.868 $\pm$ 0.023 & 0.977 $\pm$ 0.015  \\
POS\tablefootnote{Encoded parts of speech features (section \ref{subsection:parts-of-speech-feat})}    & BART-large    & 0.906 $\pm$ 0.011 & 0.987 $\pm$ 0.005 \\
Tf-idf & DeBERTa-large & \textbf{0.911 $\pm$ 0.005} & \textbf{0.987 $\pm$ 0.008} \\ 
\hline
\end{tabular}}
\caption{Results with Additional Features}
\label{tab:add_features_table}
\end{table}

Table \ref{tab:additional datasets table} shows the outcome of training on additional datasets such as the STS and the IBM Rank 30k dataset without using any feature vectors. We find that the best performing scores using both these datasets are almost equal and are achieved by the same BART-large model architecture. Thus training on additional datasets led to a substantial increase in both \emph{mAP} strict and \emph{mAP} relaxed scores. The best results of pre-training on the additional datasets were almost similar, which might be because the ground truth scores in both the datasets effectively reflect the semantic overlap between two sentences (i.e., if two sentences of a data sample are semantically similar, they would have a higher score, and vice versa), thus making the datasets similar to one another. 

We also tried adding feature vectors plus training on additional datasets\footnote{The results of these experiments can be found in Appendix available in the supplementary material.}, but there was no significant change in the performance than the existing results. Transformers themselves are able to learn syntactic and semantic features on their own during the training process \citep{clark2019does}. Adding these features only increases redundancy, as a result of which the performance of the model isn't affected much. This observation could also be seen in the difference in the results of table \ref{tab: table of vanilla text results} and \ref{tab:add_features_table}.

Complete results of these experiments can be found in the Appendix available in the supplementary material.


\begin{table}[h!]
\centering
\resizebox{\columnwidth}{!}{\begin{tabular}{llll}
\hline
\textbf{Model} &
  \textbf{\begin{tabular}[c]{@{}l@{}}Additional \\ Dataset\end{tabular}} &
  \textbf{\begin{tabular}[c]{@{}l@{}}mAP \\ Strict\end{tabular}} &
  \textbf{\begin{tabular}[c]{@{}l@{}}mAP \\ Relaxed\end{tabular}} \\
\hline
BERT-large    & STS     & 0.818 $\pm$ 0.045          & 0.933 $\pm$ 0.016          \\
RoBERTa-large & STS     & 0.905 $\pm$ 0.007          & \textbf{0.986 $\pm$ 0.004} \\
BART-large    & STS     & \textbf{0.920 $\pm$ 0.005} & 0.967 $\pm$ 0.036          \\
DeBERTa-large & STS     & 0.912 $\pm$ 0.004         & 0.983 $\pm$ 0.003          \\
\hline
BERT-large    & IBM Rank 30k & 0.793 $\pm$ 0.029          & 0.914 $\pm$ 0.019          \\
RoBERTa-large & IBM Rank 30k & 0.872 $\pm$ 0.006          & 0.974 $\pm$ 0.003          \\
BART-large    & IBM Rank 30k & \textbf{0.921 $\pm$ 0.018} & \textbf{0.982 $\pm$ 0.002} \\
DeBERTa-large & IBM Rank 30k & 0.894 $\pm$ 0.017          & 0.982 $\pm$ 0.008         \\
\hline
\end{tabular}}
\caption{Results with pretraining on additional datasets}
\label{tab:additional datasets table}
\end{table}

\begin{figure}[h!]
  \includegraphics[width=\linewidth]{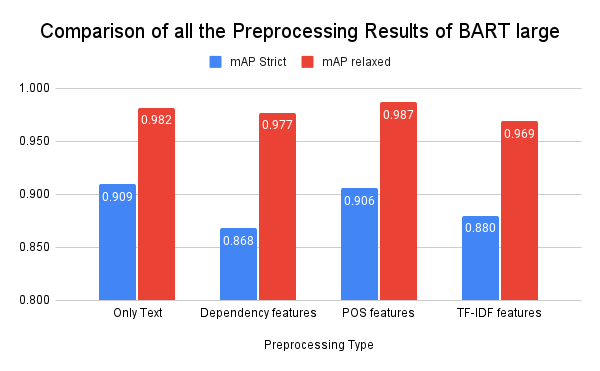}
  \caption{All preprocessing methods with BART large}
  \label{fig:bartlarge}
\end{figure}

\begin{figure}[h!]
  \includegraphics[width=\linewidth]{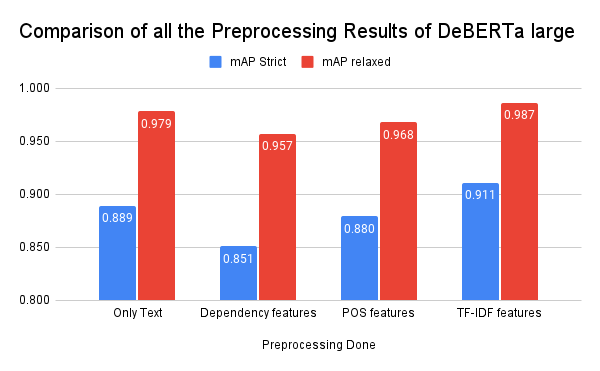}
  \caption{All preprocessing methods with DeBERTa large}
  \label{fig:debertalarge}
\end{figure}

In Figures \ref{fig:bartlarge} and \ref{fig:debertalarge}, we plot the results of the best-performing transformer-based models using different feature vectors.

 

\section{Ablation Study}

We designed different settings to compare and validate our approach and its performance. This section consists of results on excluding of topics from input in Section~\ref{subsection: exclude-topic}, incorporating average of hidden states before feeding to dense layers in Section~\ref{subsection: average-hidden-states}, and boosting in Section~\ref{subsection: boosting}. Since we obtain best results with BART-large and DeBERTa-large with Tf-idf features, thus the following ablation study is done with these class of models.

\subsection{Exclusion of topic from input}
\label{subsection: exclude-topic}

We incorporate the combination of keypoints and arguments as input to the pre-trained language models to analyze the importance of the topic towards generating the matching score. Comparing Table \ref{tab: table of vanilla text results} and Table \ref{tab:exclusion of topic}, incorporating topic provides more context in the input, thus improving both \emph{mAP} strict score and \emph{mAP} relaxed score. 

\begin{table}[htbp]
\centering
\resizebox{\columnwidth}{!}{\begin{tabular}{ccc}
\hline
\textbf{Model} & \multicolumn{1}{l}{\textbf{mAP Strict}} & \multicolumn{1}{l}{\textbf{mAP Relaxed}} \\
\hline
BART-base     & 0.803 $\pm$ 0.028          & 0.898 $\pm$ 0.015          \\
DeBERTa-base  & 0.823 $\pm$ 0.030          & 0.922 $\pm$ 0.012          \\
BART-large    & \textbf{0.880 $\pm$ 0.006} & \textbf{0.946 $\pm$ 0.010} \\
DeBERTa-large & 0.874 $\pm$ 0.025          & 0.946 $\pm$ 0.027        \\
\hline
\end{tabular}}
\caption{Results with input as keypoint plus argument}
\label{tab:exclusion of topic}
\end{table}

\subsection{Average of hidden states}
\label{subsection: average-hidden-states}
We average the last two and the last three hidden states of the pre-trained language model. The average hidden states were then fed into the dense layers to obtain the match score. It can be observed that for both BART-large and DeBERTa-large, the performance decreases as we incorporate more hidden states for the output. The intuition behind this observation can be attributed to the fact that task-specific information encoded in hidden states is less as compared to the last layer, resulting in decreased performance. 
The results are shown in Table \ref{tab:ablation-hidden-average}.

\begin{table}[htbp]
\centering
\resizebox{\columnwidth}{!}{\begin{tabular}{cccc}
\hline
\textbf{Model} &
  \multicolumn{1}{l}{\textbf{\begin{tabular}[c]{@{}l@{}}No. of \\ Hidden States\end{tabular}}} &
  \multicolumn{1}{l}{\textbf{mAP Strict}} &
  \multicolumn{1}{l}{\textbf{mAP Relaxed}} \\
\hline
BART-large    & 2 & 0.868 $\pm$ 0.016 & 0.941 $\pm$ 0.004 \\
DeBERTa-large & 2 & \textbf{0.871 $\pm$ 0.039}     & \textbf{0.949 $\pm$  0.015}          \\
BART-large    & 3 & 0.837 $\pm$ 0.020          & 0.933 $\pm$ 0.012          \\
DeBERTa-large & 3 & 0.850 $\pm$ 0.014          & 0.934 $\pm$ 0.022       \\
\hline
\end{tabular}}
\caption{Results with average of hidden states}
\label{tab:ablation-hidden-average}
\end{table}

\subsection{Boosting}
\label{subsection: boosting}

We implemented the AdaBoost algorithm by considering our baseline transformer architecture as the base model for this sequential paradigm. BART-large and DeBERTa-large were the transformers used for this study. The first base model was trained with the whole training set, whereas the other four models were trained by sampling data points from a probability distribution. 
Initially, all the data points were assigned an equal probability. However, the distribution was updated in a way such that the erroneous data points for the previous base models were given a higher probability to be sampled. 

The top $10,000$ most probable data points were sampled for each base model except for the first one. It can be observed from Table \ref{tab: table of vanilla text results} and Table \ref{tab:ablation-boosting} that for DeBERTa large model, the \emph{mAP} Strict has indeed been boosted from $0.889$ to $0.904$. The results are mentioned in Table \ref{tab:ablation-boosting}.

\begin{table}[h!]
\centering
\resizebox{\columnwidth}{!}{\begin{tabular}{ccc}
\hline
\textbf{Model} & \multicolumn{1}{l}{\textbf{mAP Strict}} & \multicolumn{1}{l}{\textbf{mAP Relaxed}} \\
\hline
BART-large     & 0.832 $\pm$ 0.020                          & 0.960 $\pm$ 0.010                           \\
DeBERTa-large  & \textbf{0.904 $\pm$ 0.021}  & \textbf{0.973 $\pm$ 0.017}   \\
\hline
\end{tabular}}
\caption{Boosting Results on Transformer model}
\label{tab:ablation-boosting}
\end{table}

\section{Conclusion} 
\label{sec:conclusion}
\vspace{-0.2cm}

In this work, we used Pre-trained Language Models (PLMs) to predict the match score for each argument and keypoint pair under the same stance towards the topic. We observed the state-of-the-art PLMs such as BART and DeBERTa perform the best compared to other models. We further improve the performance with additional datasets (IBM Rank 30k and STS) to perform additional pre-training (with sentence similarity) before fine-tuning on ArgKP-2021 dataset. We experimented with POS, Dependency and Tf-idf features to evaluate the addition of extra syntactic features. We support the selection of our final models with various ablation studies. It would be a good future  direction to generate appropriate explanations from concatenated input and propose methods to use explanations in the training process.

\section{Acknowledgements} 
\label{sec:acknow}
\vspace{-0.2cm}

We would like to thank the organizers Roni Friedman-Melamed, Lena Dankin, Yufang Hou, and Noam Slonim for holding this shared task. It was a great learning experience for us. We would also like to thank our fellow participants at ArgMining 2021; we look forward to learning more about their approaches and interacting with them at EMNLP. Finally, we would like to extend a big thanks to makers and maintainers of the exemplary HuggingFace \citep{wolf2020huggingfaces} repository, without which most of our research would have been impossible.

\bibliography{anthology,custom}
\bibliographystyle{acl_natbib}
\end{document}